\documentclass[runningheads]{llncs}

\usepackage[T1]{fontenc}
\usepackage[utf8]{inputenc}

\usepackage{graphicx}
\usepackage[table,dvipsnames]{xcolor} 

\usepackage{amsfonts}
\usepackage{nicefrac}
\usepackage{enumitem}
\usepackage{amsmath}
\usepackage[protrusion=true,expansion=false]{microtype}
\usepackage{booktabs}   
\usepackage{tabularx}   
\usepackage{multirow}   
\usepackage{float}      
\usepackage{tablefootnote} 
\usepackage{wrapfig}
\usepackage{hyperref}
\usepackage{url}
\usepackage{bookmark}
\usepackage{marvosym}

\title{
DialectS2S: End-to-End Speech Dialogue Modeling for Low-Resource Chinese Dialects
}
\titlerunning{DialectS2S for Low-Resource Chinese Dialects}

\author{
Yi Shu \inst{1} \and Tianyu Peng\inst{1,2,3} \and Yingzhuo Deng\inst{1,2}\and Wen Yang\inst{1,2} \and
Jun Lin\inst{4}\and Changming Xie\inst{4}\and Xinyu Yu\inst{4} \and Jiajun Zhang\inst{1,2,3}\textsuperscript{(\Letter)}}
\institute{
School of Artificial Intelligence, University of Chinese Academy of Sciences 
\and Institute of Automation, Chinese Academy of Sciences 
\and Wuhan AI Research 
\and GWM AI Lab\\
\email{shuyi23@mails.ucas.ac.cn}
\email{jjzhang@nlpr.ia.ac.cn}
}
\authorrunning{Y. Shu et al.} 
\begin{document}
\maketitle
\begingroup
\renewcommand{\thefootnote}{(\Letter)}
\footnotetext{Corresponding author}
\endgroup
\begingroup
\renewcommand{\thefootnote}{}
\footnotetext{Resources: \href{https://huggingface.co/CASIA-LM/DialectS2S}{model checkpoints}, \href{https://huggingface.co/datasets/CASIA-LM/DialectS2S_Datasets}{training datasets}, and \href{https://github.com/CASIA-LM/OpenS2S}{fine-tuning code}.}
\endgroup
\begin{abstract}

Current end-to-end speech dialogue models are primarily optimized for mainstream languages and remain limited in low-resource dialect scenarios due to the scarcity of dialect speech data. Moreover, during dialect adaptation, the semantic representation space of speech dialogue models continuously evolves, while conventional speech supervision remains unchanged, leading to semantic inconsistency between hidden representations and speech targets and degrading speech stability and naturalness. To address these issues, we propose DialectS2S, an end-to-end speech dialogue model for Chinese dialects. We first develop a scalable dialect speech dialogue synthesis pipeline for efficient data construction. We further introduce a two-stage post-training strategy with self-aligned speech supervision, which aligns the semantic content of speech supervision with the evolved semantic representations of the model to improve dialect speech generation quality. Experimental results show that DialectS2S consistently outperforms existing baselines across multiple Chinese dialects in speech dialogue, achieving substantial improvements in dialect consistency, response quality, and speech intelligibility. Our work provides an efficient and scalable solution for end-to-end speech dialogue modeling in low-resource dialect scenarios. To facilitate future research and practical applications, we fully open-source the DialectS2S framework, including model checkpoints, training datasets, and fine-tuning code.

\end{abstract}

\section{Introduction}

Dialects play an important role in regional communication and cultural expression. Recent end-to-end speech dialogue models~\cite{zeng2024glm,xu2025qwen2,ding2025kimi,wu2025step,wang2025opens2s,cui2026minicpm} have achieved strong conversational performance in mainstream languages such as Mandarin Chinese and English. However, support for low-resource Chinese dialects remains limited. Existing Chinese dialect speech research mainly focuses on speech recognition or speech synthesis, while unified end-to-end speech dialogue modeling for Chinese dialects remains largely underexplored.

Current speech dialogue models often rely on large-scale speech dialogue corpora, while high-quality dialect speech data remain scarce. Moreover, recent studies have shown that directly supervised fine-tuning speech dialogue models may degrade speech quality and generation stability~\cite{zhang2025echox}. This issue becomes more challenging in low-resource dialect scenarios due to limited high-quality supervision and large acoustic variations. Therefore, enabling stable and natural dialect speech interaction under limited supervision remains a major challenge for end-to-end speech dialogue systems.

To address these challenges, we propose DialectS2S, an end-to-end speech dialogue model for low-resource Chinese dialects. DialectS2S supports multilingual interaction across five languages and dialects, including Mandarin, English, Sichuanese, Cantonese, and the Tianjin dialect. To construct dialect speech dialogue data, we develop a scalable synthesis pipeline that rewrites existing dialogue texts into dialectal expressions using large language models and synthesizes corresponding dialect speech with speech generation models. Furthermore, we introduce a two-stage post-training strategy with self-aligned speech supervision to improve speech generation stability and naturalness during dialect adaptation.

In summary, our main contributions are as follows:

\begin{enumerate}[label=(\arabic*)]

\item \textbf{End-to-End Speech Dialogue Model for Chinese Dialects:} We propose DialectS2S, an end-to-end speech dialogue model for low-resource Chinese dialects. To facilitate future research on dialect speech interaction, we fully open-source the DialectS2S framework, including model checkpoints, training datasets, and fine-tuning code.

\item \textbf{Self-Aligned Speech Supervision for Dialect Speech Generation:} We propose a two-stage post-training strategy with self-aligned speech supervision, which aligns speech targets with the model's semantic predictions and improves speech intelligibility while preserving response quality.

\item \textbf{Strong Multi-Dialect Performance:} Experimental results demonstrate that DialectS2S significantly outperforms existing open-source baselines in dialect consistency, response quality, and speech intelligibility across multiple Chinese dialects.

\end{enumerate}

\section{Related Work}
\label{others}

\subsection{End-to-End Large Speech Models}

Recent end-to-end large speech models have achieved strong performance in speech recognition, synthesis, and dialogue tasks. Compared with cascaded systems, these models directly map speech inputs to text and speech outputs, reducing error accumulation and improving interaction naturalness. LLaMA-Omni~\cite{fang2025llama} appends a speech decoder to a large language model for unified speech understanding and generation. GLM-4-Voice~\cite{zeng2024glm} adopts interleaved text-speech modeling for streaming interaction. Step-Audio 2~\cite{wu2025step} and Step-Audio-R1~\cite{tian2025step} follow a similar interleaved architecture. Moshi~\cite{defossez2024moshi}, Voila~\cite{shi2025voila} and MiniCPM-o~\cite{cui2026minicpm} further support low-latency full-duplex conversations. More recently, Qwen2.5-Omni~\cite{xu2025qwen2} and Qwen3-Omni~\cite{xu2025qwen3} adopt the Thinker-Talker architecture, where a large language model handles semantic reasoning and a smaller model generates speech. OpenS2S~\cite{wang2025opens2s} further improves fine-grained speech understanding and releases open-source models and training pipelines. Despite these advances, existing systems are mainly optimized for high-resource languages such as Mandarin and English, while low-resource dialect speech interaction remains underexplored.

\subsection{Low-Resource Dialect Speech Modeling}

Compared with high-resource languages such as English and Mandarin Chinese, speech modeling for Chinese dialect varieties faces challenges including limited data availability, expensive annotation, and substantial pronunciation variation. Existing studies mainly focus on dialect speech corpora, automatic speech recognition (ASR), and text-to-speech synthesis (TTS). KeSpeech~\cite{tang2021kespeech} constructs a large-scale corpus covering eight Chinese dialects and more than 27,000 speakers, while WenetSpeech-Yue~\cite{Li2026} and WenetSpeech-Chuan~\cite{dai2026wenetspeech} provide high-quality Cantonese and Sichuanese speech datasets. Systems such as FunASR~\cite{an2025fun} and FireRedASR2S~\cite{xu2026fireredasr2s} support multilingual and multi-dialect speech processing. In TTS, DIAMOE-TTS~\cite{chen2025diamoe} enables unified dialect modeling through IPA representations, while CosyVoice2~\cite{du2024cosyvoice} improves speech naturalness and speaker consistency and supports dialect speech generation from reference audio. However, existing work primarily addresses speech recognition or synthesis in isolation, with limited exploration of end-to-end spoken dialogue systems for Chinese dialects.

\section{Method}

\subsection{Overview}

We propose an efficient framework for low-resource dialect adaptation in end-to-end speech dialogue models. The framework consists of a dialect speech dialogue synthesis pipeline and a two-stage post-training procedure, including mixed-data supervised fine-tuning and self-aligned speech supervision training. Through the proposed framework, pretrained speech dialogue models can acquire stable dialect understanding and generation capabilities while preserving their original multilingual ability.

\subsection{Dialect Speech Dialogue Synthesis Pipeline}

The proposed synthesis pipeline consists of three stages: dialect text generation, dialect speech synthesis, and speech naturalness filtering.

\paragraph{\textbf{Dialect Dialogue Generation}}

Directly prompting large language models to generate dialect dialogues often produces repetitive and low-diversity content. Instead, we rewrite the dialogue portion of an open-source Mandarin dialogue dataset~\cite{wang2025opens2s} into dialectal expressions using large language models~\cite{yang2025qwen3} while preserving the original semantics and dialogue structure. This strategy efficiently reuses existing dialogue corpora and enables low-cost construction of diverse dialect dialogue texts.

\paragraph{\textbf{Dialect Speech Synthesis}}

We adopt different synthesis strategies for user-query speech and system-response speech. For user-query speech, we select 100 dialect seed utterances for each dialect from open-source dialect datasets~\cite{tang2021kespeech,Li2026,dai2026wenetspeech}, including both male and female speakers. These utterances are used as reference audio for CosyVoice2~\cite{du2024cosyvoice} to synthesize dialect query speech from the rewritten dialogue texts, ensuring diverse vocal characteristics in the input speech. For system-response speech, we adopt a consistent speaker timbre strategy for response synthesis. Specifically, we first construct dialect-specific reference speech conditioned on a target speaker and then use the generated reference speech for response synthesis. This strategy maintains consistent voice characteristics in synthesized responses and further improves dialect expressiveness and speech naturalness.

\paragraph{\textbf{Speech Naturalness Filtering}}

To improve data quality, we further introduce a speech filtering stage using UTMOS~\cite{saeki22c_interspeech} to automatically remove synthesized samples with unclear pronunciation or unnatural prosody.

\subsection{Model Training}

\begin{figure}[htbp]
  \centering
  \includegraphics[width=0.8\textwidth]{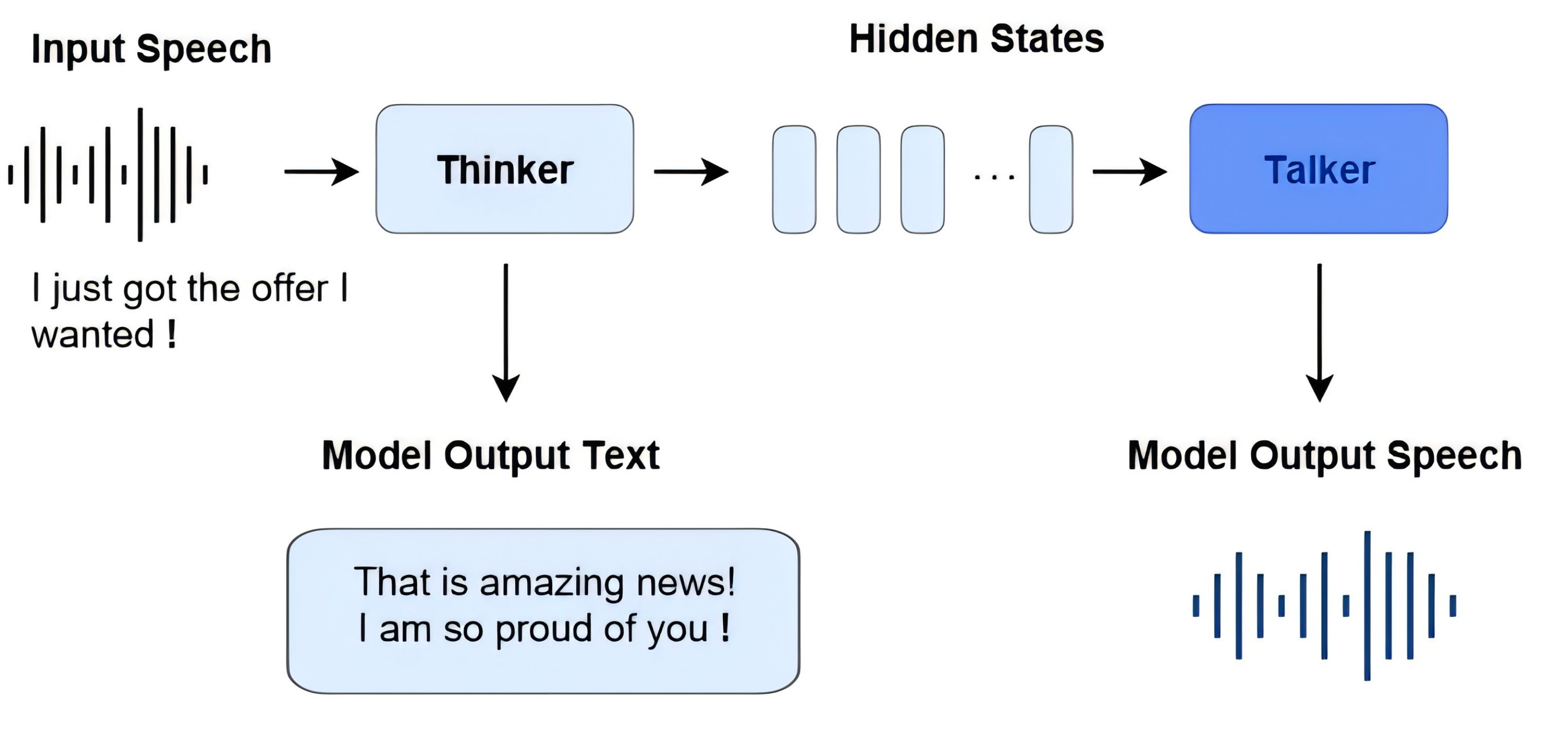}
  \caption{The proposed unified Thinker-Talker architecture.}
  \label{fig:framework}
\end{figure}

Most existing end-to-end speech-to-speech dialogue models~\cite{zeng2024glm,xu2025qwen2,ding2025kimi,wu2025step,wang2025opens2s,xu2025qwen3,tian2025step} are built upon pretrained text language models and synchronously generate both text and speech responses from speech inputs. In this work, we interpret such systems from a unified Thinker-Talker perspective. Specifically, given an input speech sequence $\mathbf{X}$, the \textit{Thinker} module performs speech understanding and semantic reasoning, producing intermediate hidden representations $\mathbf{H}$ together with text response tokens $\mathbf{T}$. Conditioned on $\mathbf{H}$, the \textit{Talker} module learns the mapping from hidden semantic representations to speech response tokens $\mathbf{U}$ by leveraging the semantic and paralinguistic information encoded in the hidden states. The overall architecture is illustrated in Figure~\ref{fig:framework}.

Although these models effectively leverage pretrained language models for end-to-end speech interaction, adapting them to low-resource dialect scenarios often introduces shifts in the hidden-state distribution during post-training. Such shifts are not necessarily undesirable, since the model must gradually acquire new dialect-specific semantic and acoustic representations. However, the speech supervision used during training often remains semantically inconsistent with the evolved hidden representations. As a result, the Talker module must implicitly perform semantic recovery in addition to speech generation, increasing optimization difficulty and degrading speech quality.

\begin{wrapfigure}{r}{0.32\textwidth}
  \centering
  \vspace{-25pt}
  \includegraphics[width=0.30\textwidth]{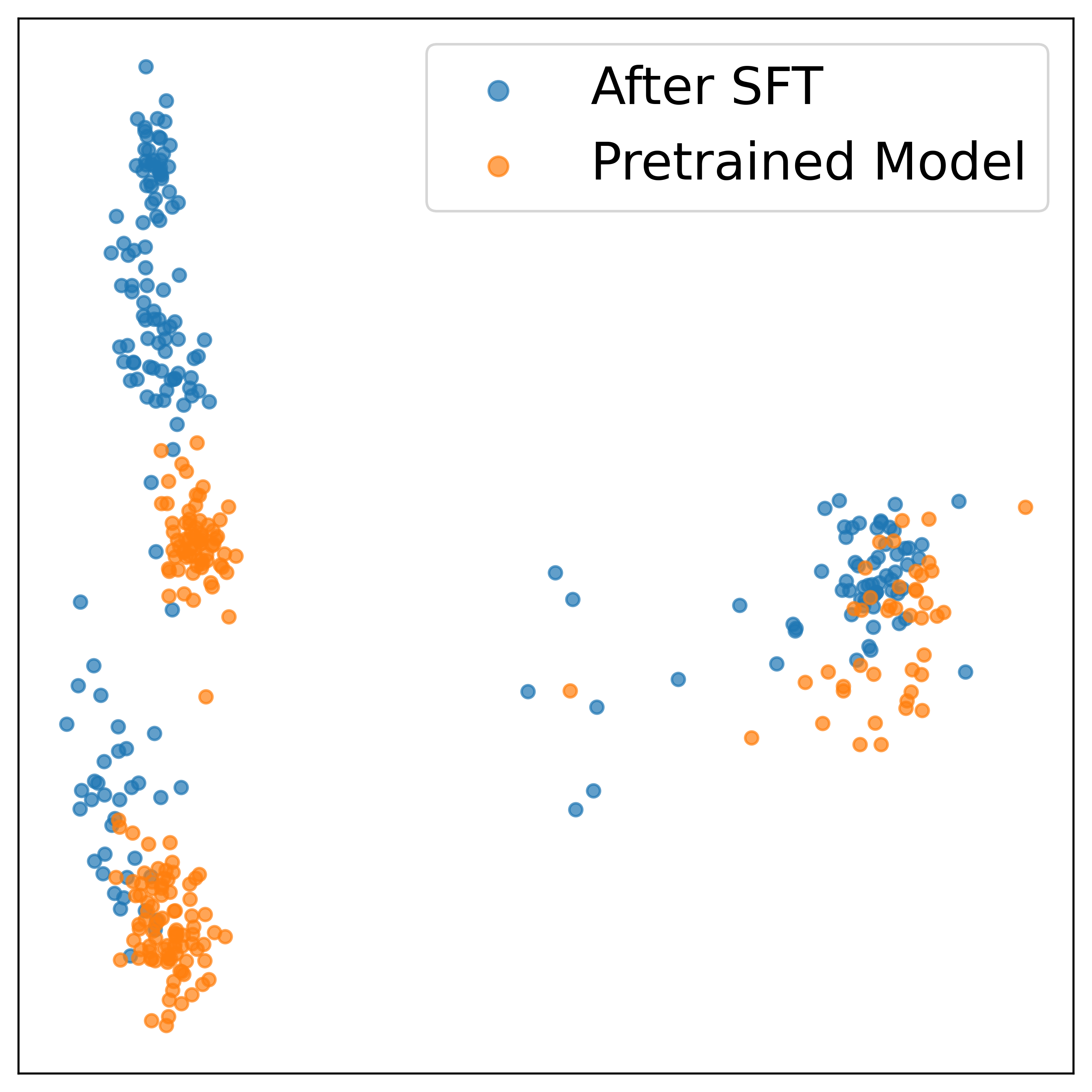}
  \vspace{-10pt}
  \caption{Visualization of the final-layer Thinker hidden states before and after supervised fine-tuning.}
  \label{fig:hidden}
  \vspace{-20pt}
\end{wrapfigure}

As illustrated in Figure~\ref{fig:hidden}, we visualize the hidden-state distributions using PCA projection on the token-averaged final-layer Thinker representations. The hidden representations continuously evolve after supervised fine-tuning, highlighting the necessity of adapting speech supervision to the updated semantic space.

To better align speech supervision with the evolving semantic representations during dialect adaptation, we adopt a two-stage training framework consisting of \textbf{Mixed-Data SFT} and \textbf{Self-aligned Speech Supervision Training}. The first stage enables the model to acquire dialect-aware semantic representations through supervised fine-tuning, while the second stage further aligns speech generation with the evolved hidden representations to improve dialect speech quality and intelligibility.

\paragraph{\textbf{Mixed-Data SFT}}

Initialized from a speech dialogue model~\cite{wang2025opens2s} with strong conversational performance in mainstream languages, we perform full-parameter supervised fine-tuning on mixed dialect, Mandarin, and English speech dialogue data. The training objective is defined as:

\begin{equation}
\label{eq:dialects2s_sft}
\mathcal{L}_{SFT}
=
\underbrace{
-\sum_{i=1}^{L}
\log P(t_i \mid \mathbf{X}, \mathbf{T}_{<i}; \theta_{Thinker})
}_{\text{Thinker Loss}}
-
\underbrace{
\sum_{j=1}^{V}
\log P(u_j \mid \mathbf{H}, \mathbf{U}_{<j}; \theta_{Talker})
}_{\text{Talker Loss}}
\end{equation}

where $\theta_{Thinker}$ and $\theta_{Talker}$ denote the parameters of the Thinker and Talker modules.

This stage gradually shifts the hidden representations of the Thinker module toward dialect-aware semantic spaces through supervised fine-tuning, enabling the model to acquire initial dialect interaction capabilities. However, the Talker module still struggles to fully adapt to the evolved semantic representations, since the representation space of the Thinker module continues to evolve during training while the speech supervision remains unchanged.

\paragraph{\textbf{Self-aligned Speech Supervision Training}}

To better align speech supervision with the evolved semantic representations of the Thinker module, we further propose self-aligned speech supervision training.

We first use the fine-tuned Thinker module to generate text predictions:

\begin{equation}
\hat{\mathbf{T}}
=
\mathrm{Thinker}(\mathbf{X})
\end{equation}

Aligned speech supervision is then synthesized from the predicted text using a dialect TTS model:

\begin{equation}
\hat{\mathbf{U}}
=
\mathrm{TTS}(\hat{\mathbf{T}})
\end{equation}

The synthesized speech is subsequently used as the new supervision target, such that the semantic content of the speech supervision is explicitly aligned with the semantic information encoded in the hidden representations for Talker modeling. To maintain stable coordination between the Thinker and Talker modules, the self-aligned stage still adopts end-to-end full-parameter optimization, although the self-aligned loss is primarily applied to the Talker module.

\begin{equation}
\label{eq:self_align}
\mathcal{L}_{Align\_Talker}
=
-
\sum_{j=1}^{V}
\log
P(
\hat{u}_j
\mid
\mathbf{H},
\hat{\mathbf{U}}_{<j};
\theta_{Talker}
)
\end{equation}

\begin{figure}[htbp]
  \centering
  \includegraphics[width=1\textwidth]{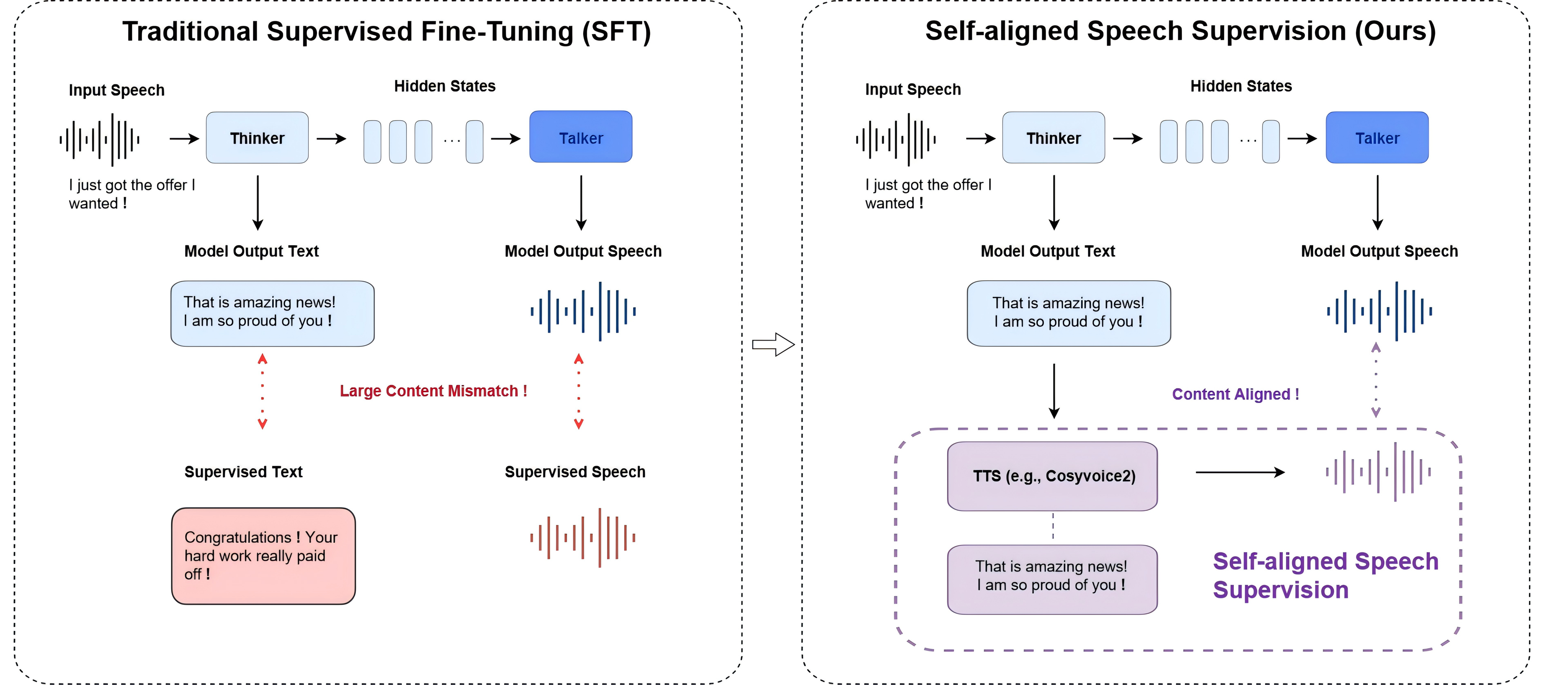}
  \caption{Self-aligned supervision aligns speech targets with semantic representations produced by the Thinker module, enabling the Talker module to focus on prosody and pronunciation modeling.}
  \label{fig:modelwork}
\end{figure}

By aligning speech supervision with the model's semantic predictions, the proposed strategy reduces semantic mismatch between hidden representations and speech targets, allowing the Talker module to focus on speech generation rather than semantic correction. Figure~\ref{fig:modelwork} provides a comparison between conventional supervised fine-tuning and the proposed self-aligned speech supervision strategy.

\section{Experiments}
\label{headings}

\subsection{Datasets}

Based on a high-quality open-source speech dialogue dataset~\cite{wang2025opens2s}, we construct dialect speech dialogue corpora for three Chinese dialects, including Sichuanese, Cantonese, and the Tianjin dialect, using the proposed synthesis pipeline. Each dialect contains approximately 3,600 speech query-response pairs. In addition, 3,600 high-MOS speech dialogue samples are selected for both Mandarin and English from this dataset and mixed with the constructed dialect data for joint training. The final training set contains approximately 18,000 speech dialogue pairs with a total duration of 225 hours, primarily consisting of open-domain daily conversations to better reflect real-world multilingual and dialect interaction scenarios.

\subsection{Training Setup}

We initialize the model from OpenS2S~\cite{wang2025opens2s} due to its fully open-source nature and strong conversational capabilities in everyday spoken interactions. We first perform two epochs of supervised fine-tuning to obtain a base model with stable dialect capabilities, followed by self-aligned speech supervision training on the same dataset. All experiments are conducted on 8 NVIDIA A800 GPUs using DeepSpeed~\cite{rasley2020deepspeed} with ZeRO Stage-2~\cite{rajbhandari2020zero}. The model is optimized using AdamW~\cite{loshchilov2017decoupled} with a learning rate of $2 \times 10^{-5}$, $\beta_1=0.9$, $\beta_2=0.999$, $\epsilon=1\times10^{-8}$, and weight decay 0.05. The learning rate schedule follows WarmupDecayLR.

\section{Evaluation}

\subsection{Evaluation Setup}

We evaluate DialectS2S on multilingual speech interaction from three aspects: language matching accuracy, response quality, and speech intelligibility. A multilingual dialect speech dialogue benchmark containing 250 test samples is constructed, covering two mainstream languages and three Chinese dialects, with evaluation queries built using different reference audios and query texts. Comparisons are conducted against several representative open-source speech dialogue models, including GLM-4-Voice~\cite{zeng2024glm}, Qwen2.5-Omni~\cite{xu2025qwen2}, Kimi-Audio~\cite{ding2025kimi}, Step-Audio2~\cite{wu2025step}, OpenS2S~\cite{wang2025opens2s}, and MiniCPM-o-4.5~\cite{cui2026minicpm}.

\subsection{Response Language Matching Accuracy}

We first evaluate whether the model can generate speech responses in the input language or dialect while preserving multilingual capability. Specifically, we use FireRedLID~\cite{xu2026fireredasr2s} to identify the language of generated speech responses. All evaluations are conducted in a zero-shot setting without text prompts or explicit language control signals. Language matching accuracy is computed as the proportion of valid samples whose predicted language matches the input language. Results are shown in Table~\ref{tab:language_match}. DialectS2S achieves substantial improvements across all dialect scenarios while maintaining strong Mandarin and English performance, significantly outperforming existing open-source baselines.

\begin{table}[H]
  \centering
  \caption{Comparison of language matching accuracy (\%) between DialectS2S and baseline models.}
  \label{tab:language_match}
  \renewcommand{\arraystretch}{1}

  \begin{tabularx}{\textwidth}{l|*{6}{>{\centering\arraybackslash}X}}
      \toprule[1.2pt]
      \multirow{2}{*}{\textbf{Model}} & \multicolumn{6}{c}{\textbf{Language Matching Accuracy}} \\
      \cline{2-7}
      & English & Mandarin & Sichuanese & Cantonese & Tianjin & Avg.\\ 
      \midrule[0.8pt]

      GLM-4-Voice & 94.00 & 97.92 & 2.22 & 46.94 & 0.00 & 48.22\\
      Qwen2.5-Omni & 100.00 & 100.00 & 0.00 & 0.00 & 0.00 & 40.00\\
      Kimi-Audio  & 72.00 & 100.00 & 16.00 & 0.00 & 0.00 & 37.60\\
      Step-Audio2 & 100.00 & 86.00 & 0.00 & 10.00 & 0.00 & 39.20\\
      OpenS2S     & 100.00 & 93.88 & 0.00 & 0.00 & 0.00 & 38.78\\
      MiniCPM-o-4.5 & 100.00 & 98.00 & 2.04 & 52.21 & 2.00 & 50.85\\
      \textbf{DialectS2S} & 100.00 & 97.96 & \textbf{96.00} & \textbf{95.00} & \textbf{91.00} & \textbf{95.99} \\

      \bottomrule[1.2pt]
  \end{tabularx}
\end{table}

\subsection{Response Quality}

Response quality is further evaluated on the constructed multilingual dialect benchmark using the text responses synchronously generated during speech interaction. Qwen3-Plus~\cite{yang2025qwen3} is employed as the automatic evaluator to assess naturalness, semantic accuracy, and conformity to dialectal language usage, with scores ranging from 1 to 5. The scoring rubric is provided in Appendix~\ref{app:details}. Since all compared models support simultaneous text and speech generation, no additional ASR system is required. As shown in Table~\ref{tab:content_quality}, DialectS2S achieves the highest overall average score and consistently delivers superior response quality across dialect scenarios, demonstrating strong semantic understanding and dialect-aware response generation capability.

\begin{table}[H]
  \centering
  \caption{Response quality evaluation across different language and dialect scenarios.}
  \label{tab:content_quality}

  \renewcommand{\arraystretch}{1}

  \begin{tabularx}{\textwidth}{l|*{6}{>{\centering\arraybackslash}X}}
      \toprule[1.2pt]
      \multirow{2}{*}{\textbf{Model}} & \multicolumn{6}{c}{\textbf{Response Quality}} \\
      \cline{2-7}

      & English & Mandarin & Sichuanese & Cantonese & Tianjin & Avg. \\
      \midrule[0.8pt]

      GLM-4-Voice & 4.56 & 4.40 & 3.10 & 4.17 & 2.19 & 3.68 \\
      Qwen2.5-Omni & 4.60 & 4.65 & 3.12 & 3.26 & 3.11 & 3.75 \\
      Kimi-Audio  & 4.07 & 4.12 & 3.22 & 3.80 & 2.26 & 3.49 \\
      Step-Audio2 & 4.20 & 3.92 & 2.78 & 3.02 & 2.06 & 3.20 \\
      OpenS2S     & 4.54 & 4.36 & 2.96 & 3.26 & 2.60 & 3.54 \\
      MiniCPM-o-4.5 & 4.22 & 4.45 & 3.51 & 4.24 & 2.30 & 3.74 \\
      \textbf{DialectS2S}  & 4.60 & 4.33 & \textbf{4.48} & \textbf{4.28} & \textbf{4.40} & \textbf{4.42} \\

      \bottomrule[1.2pt]
  \end{tabularx}
\end{table}

\subsection{Speech Intelligibility}

We use character error rate (CER) to measure speech intelligibility, where lower CER indicates more intelligible speech outputs. The text responses generated by the model are used as references, while FRASR2-AED~\cite{xu2026fireredasr2s} transcribes the generated speech responses for CER computation. Previous experiments show that existing baseline models mainly generate Mandarin Chinese rather than dialect speech under dialect interaction settings. Therefore, CER evaluation is instead conducted against the average CER reported by the FRASR2-AED dialect benchmark.

\begin{wraptable}{r}{0.42\textwidth}
  \vspace{-30pt}
  \centering
  \small
  \caption{CER (\%) of generated speech compared with the average CER reported by the FRASR2-AED benchmark.}
  \label{tab:asr_cer}
  \vspace{5pt}
  \begin{tabular}{lcc}
  \toprule
  & Mandarin & Dialect \\
  \midrule
  DialectS2S & 3.28 & \textbf{8.69} \\
  Benchmark Avg. & 3.05 & 11.67 \\
  \bottomrule
  \end{tabular}
  \vspace{-20pt}
  \end{wraptable}

As shown in Table~\ref{tab:asr_cer}, DialectS2S achieves low CER across both Mandarin and dialect scenarios, demonstrating strong speech intelligibility. In particular, DialectS2S achieves lower CER than the average performance reported by the FRASR2-AED dialect benchmark in dialect settings, indicating highly intelligible generated dialect speech.

\section{Ablation Study}

To evaluate the effectiveness of the proposed self-aligned speech supervision strategy, we conduct ablation studies under three training configurations: \textbf{sft-2ep}, which applies two epochs of supervised fine-tuning; \textbf{sft-3ep}, which continues supervised fine-tuning for one additional epoch; and \textbf{DialectS2S}, which applies self-aligned speech supervision on top of the sft-2ep model. Notably, sft-3ep and DialectS2S are trained with comparable numbers of optimization steps, enabling a fair comparison between continued conventional fine-tuning and the proposed training strategy.

We evaluate both response quality and speech intelligibility, with results reported in Table~\ref{tab:ablation_quality} and Table~\ref{tab:ablation_cer}, respectively. As shown in Table~\ref{tab:ablation_quality}, DialectS2S achieves response quality comparable to sft-3ep, indicating that self-aligned speech supervision does not compromise semantic generation capability. In contrast, Table~\ref{tab:ablation_cer} shows that DialectS2S consistently outperforms both sft-2ep and sft-3ep in speech intelligibility. In particular, simply extending supervised fine-tuning does not yield consistent improvements in CER and may even lead to degradation in certain dialect settings.

These results suggest that increasing the amount of supervised fine-tuning alone is insufficient to address the hidden-state distribution shift in the Thinker-Talker architecture. In comparison, the proposed self-aligned speech supervision explicitly aligns speech targets with the model's semantic predictions, thereby reducing semantic mismatch and improving speech clarity and stability.

\begin{table}[H]
  \centering
  \caption{Response quality evaluation under different training strategies across different language and dialect scenarios. }
  \label{tab:ablation_quality}

  \renewcommand{\arraystretch}{1}

  \begin{tabularx}{\textwidth}{l|*{6}{>{\centering\arraybackslash}X}}
      \toprule[1.2pt]

      \multirow{2}{*}{\textbf{Model}} & \multicolumn{6}{c}{\textbf{Response Quality}} \\

      \cline{2-7}

      & English & Mandarin & Sichuanese & Cantonese & Tianjin & Avg. \\

      \midrule[0.8pt]

      sft-2ep & 3.97 & 4.19 & 4.02 & 3.81 & 4.36 & 4.07 \\

      sft-3ep & 4.27 & 4.40 & 4.67 & 4.19 & 4.57 & \textbf{4.42} \\

      \textbf{DialectS2S}& 4.60 & 4.33 & 4.48 & 4.28 & 4.40 & \textbf{4.42} \\
      \bottomrule[1.2pt]
  \end{tabularx}
\end{table}

\begin{table}[H]
  \centering
  \caption{Comparison of CER (\%) under different training strategies.}
  \label{tab:ablation_cer}

  \renewcommand{\arraystretch}{1}

  \begin{tabularx}{\textwidth}{l|*{6}{>{\centering\arraybackslash}X}}

      \toprule[1.2pt]

      \multirow{2}{*}{\textbf{Model}} & \multicolumn{6}{c}{\textbf{ASR CER $\downarrow$}} \\

      \cline{2-7}

      & English & Mandarin & Sichuanese & Cantonese & Tianjin & Avg. \\

      \midrule[0.8pt]

      sft-2ep & 8.68 & 5.25 & 11.29 & 17.10 & 8.43 & 10.15 \\

      sft-3ep & 11.49 & 6.85 & 9.85 & 19.80 & 8.36 & 11.27 \\

      \textbf{DialectS2S} & 5.56 & 3.28 & 8.63 & 9.53 & 7.92 & \textbf{6.98} \\

      \bottomrule[1.2pt]
  \end{tabularx}
\end{table}

\section{Conclusion}

This work presents DialectS2S, an end-to-end speech dialogue model for low-resource Chinese dialects. By combining a scalable dialect data synthesis pipeline with a two-stage post-training strategy, DialectS2S enables stable and intelligible dialect speech interaction under limited supervision. Experimental results demonstrate strong language and dialect matching, response quality, and speech intelligibility across three Chinese dialects. To facilitate future research and practical applications, we fully open-source the DialectS2S framework, including model checkpoints, training datasets, and fine-tuning code. In future work, we plan to support more dialects and explore more effective training strategies for improving dialect authenticity and expressiveness.

\begin{credits}
\subsubsection{\ackname}
This work is supported by Beijing Natural Science Foundation L259016 and the Strategic Priority Research Program of Chinese Academy of Sciences under Grant XDA04080400.
\end{credits}

\bibliographystyle{splncs04}
\bibliography{reference}

\appendix
\section{Supplementary Details}
\label{app:details}
\begingroup

\paragraph{Response-quality rubric.}
Qwen3-Plus evaluates every model's synchronously generated text with the same prompt, without an additional ASR step. The five-point scoring rubric is:
\begin{enumerate}[label=\textbf{\arabic*:}, leftmargin=*, nosep]
  \item The response is completely irrelevant or severely fails to satisfy the user's request.
  \item The response is partially relevant but contains clear errors or missing content that make it difficult to understand.
  \item The response is generally relevant but contains some errors or unnatural expressions and does not fully satisfy the user's request.
  \item The response is of high overall quality, with only minor issues, and is generally natural and easy to understand.
  \item The response fully satisfies the user's request, is natural and fluent, is accurately expressed, and is highly consistent with the dialogue context.
\end{enumerate}
For dialect inputs, the evaluator additionally considers whether the textual dialect usage is natural and consistent with the requested dialect.

\paragraph{Comparison with a cascaded system.}

DialectS2S outperforms a cascaded system with a comparable parameter count in both response quality and inference speed. The cascade baseline combines FireRedASR2, Qwen3-8B, and CosyVoice2. Response quality is evaluated by Qwen3-Plus using the five-point rubric described above.

\begin{table}[H]
  \centering
  \caption{Supplementary cascade comparison and streaming-inference latency. Latency is measured on the first 50 Sichuanese samples using one NVIDIA A800-SXM4-80GB GPU.}
  \label{tab:supplementary_results}
  \setlength{\tabcolsep}{5pt}
  \resizebox{\textwidth}{!}{%
  \begin{tabular}{lcc@{\qquad}lrrrr}
    \toprule
    \multicolumn{3}{c}{(a) Cascade comparison} & \multicolumn{5}{c}{(b) Streaming latency (s)} \\
    \cmidrule(r){1-3}\cmidrule(l){4-8}
    Metric & Cascade & DialectS2S & Metric & Mean & Median & P90 & P95 \\
    \midrule
    Success & 49/50 & \textbf{50/50} & TTFT & 0.852 & 0.532 & 0.671 & 3.698 \\
    Text startup (s) & 1.170 & \textbf{0.852} & First audio & 5.110 & 4.879 & 5.399 & 8.759 \\
    Audio startup (s) & 8.589 & \textbf{5.110} & Full generation & 42.976 & 44.225 & 76.493 & 80.753 \\
    Quality (1--5) & 2.15 & \textbf{4.48} & & & & & \\
    \bottomrule
  \end{tabular}%
  }
\end{table}

\endgroup

\end{document}